\relax
\documentclass{article}

\usepackage[final, nonatbib]{nips_2016}

\usepackage{times}
\usepackage{helvet}
\usepackage{courier}
\usepackage{amsfonts}
\usepackage{amsmath}
\usepackage{graphicx}
\usepackage{epstopdf}
\usepackage{subcaption}
\usepackage{float}

\def\etal{\emph{et al.}}

\newcommand{\THT}{\boldsymbol{\theta}}
\newcommand{\W}{\mathbf{w}}
\newcommand{\K}{\mathbf{k}}
\newcommand{\D}{\mathcal{D}}
\newcommand{\A}{\boldsymbol{\mu}}

\begin{document}

\title{Generalized Dropout}
\author{
	Suraj Srinivas\\
	Dept. of Computational \& Data Sciences \\
	Indian Institute of Science, Bangalore \\
	\texttt{surajsrinivas@grads.cds.iisc.ac.in} \\
	\And
	R. Venkatesh Babu \\
	Dept. of Computational \& Data Sciences \\
	Indian Institute of Science, Bangalore \\
	\texttt{venky@cds.iisc.ac.in} \\
}
\maketitle

\begin{abstract}
Deep Neural Networks often require good regularizers to generalize well. Dropout is one such regularizer that is widely used among Deep Learning practitioners. Recent work has shown that Dropout can also be viewed as performing Approximate Bayesian Inference over the network parameters. In this work, we generalize this notion and introduce a rich family of regularizers which we call \emph{Generalized Dropout}. One set of methods in this family, called Dropout++, is a version of Dropout with trainable parameters. Classical Dropout emerges as a special case of this method. Another member of this family selects the width of neural network layers. Experiments show that these methods help in improving generalization performance over Dropout.
\end{abstract}

\section{Introduction}
For large-scale tasks like image classification, the general practice in recent times \cite{krizhevsky2012imagenet}  has been to train large Convolutional Neural Network (CNN) models. Even with large datasets, the risk of over-fitting runs high because of the large model size. As a result, strong regularizers are required to restrict the complexity of these models. Dropout \cite{srivastava2014dropout} is a stochastic regularizer that has been widely used in recent times. However, the rule itself was proposed as a heuristic - with the objective of reducing co-adaption among neurons. As a result, it's behaviour was (and still is) not well understood. Gal and Gharamani \cite{gal2015bayesian} showed that dropout implicitly performs approximate Bayesian inference - making it a Bayesian Neural Net.

Bayesian Neural Nets (BNNs) view parameters of a Neural Network as random variables rather than fixed unknown quantities. As a result, there exists a distribution of possible values that each parameter can take. By placing an appropriate prior over these random variables, it is possible to restrict the model's capacity and implicitly perform regularization. The theoretical attractiveness of these methods is that one can now use tools from probability theory to work with these models. What advantages do BNNs offer over plain Neural Nets? First, they inherently capture uncertainty - both in the model parameters as well as predictions. Second, they are ideal for learning from small amounts of data. Third, a Bayesian approach has the advantage of distilling complex assumptions about the model in the form of prior distributions.

Inference over BNNs is typically intractable. As a result, one often uses approximations to the posterior distribution. MCMC and Variational Inference (VI) \cite{bishop2006pattern} are two popular methods for performing these approximations. In recent times, VI has emerged as the preferred method of performing this approximation as it is scalable to large models. When using VI, it is common to assume independence of model parameters. For Neural Networks, this assumption may seem unnecessarily stringent. After all, weights in a particular filter are highly correlated to produce specific patterns (an oriented edge, for instance). However, different filters in a CNN are more-or-less independent as they compute different features. In fact, it might even be advantageous to enforce independence of different filters through VI, as they reduce \emph{co-adaptation} among features. In this work, we strive to enforce independence among features rather than weights.

The overall contributions of the paper are as follows:

\begin{itemize}
\item We derive a Bayesian approach to performing inference with neural networks. In doing so, we introduce a rich family of regularizers - Generalized Dropout (GD).
\item We perform experimental analysis with Dropout++, a set of methods under GD, to understand it's behaviour.
\item We perform experiments with Stochastic Architecture Learning, another set of methods under GD, and show that they can be used to select the width of neural networks.
\item We test Dropout++ on standard networks and show that it can be used to boost performance.
\end{itemize}

\section{Bayesian Neural Networks}
In this section, we shall formally introduce the notion of BNNs and also discuss our proposed method. Let $f(\cdot ; \W)$ denote a neural network function with parameters $\W$. For a given input $x$, the neural network produces $y = f(x ; \W)$ a probability distribution over possible labels (through softmax) for a classification problem. Given training data $\D$, the parameter vector $\W$ is updated using Bayes' Rule. 

\begin{equation}
P(\W | \D) \propto P(\D| \W) P(\W)
\label{BayesRule}
\end{equation}

After computing the posterior distribution, we perform inference on a new data point $x$ as follows. Since the neural network produces a probability distribution over labels, $P(y | x, \W) = f(x ; \W)$

\begin{equation}
P(y | x, \D) = \int_w P(y | x, \W) P(\W | \D) d\W = \int_w f(x ; \W) P(\W | \D) d\W
\label{Inference}
\end{equation}

Computing the posterior from equation \ref{BayesRule} is intractable due to the complicated neural network structure. In Variational Inference, we define another distribution $q(\W)$, called the variational distribution, which approximates the posterior $P(\W | \D)$. This distribution is used instead of the posterior in equation \ref{Inference} for inference.

One common assumption when working with the variational distribution is the mean-field assumption, which requires that

\begin{equation}
q(\W) = \prod\limits_{i} q(w_i)
\end{equation}

This states that all parameters in $\W$ are independent. Such an assumption is certainly not true, especially at the feature-level, where parameters are highly correlated. Works like those of Denil \etal \cite{denil2013predicting} explicitly show that parameters of a NN can be predicted given other parameters - hinting at the large amount of correlation present. Trying to enforce independence among such weights may end up having adverse effects.

It is very difficult to overcome the independence assumption within the framework of VI. In the next section, we introduce an approach to overcome this difficulty.

\subsection{Bayesian Neural Networks with Gates}
\begin{figure*}[!t]
\begin{subfigure}{.5\textwidth}
  \centering
\includegraphics[width=4cm]{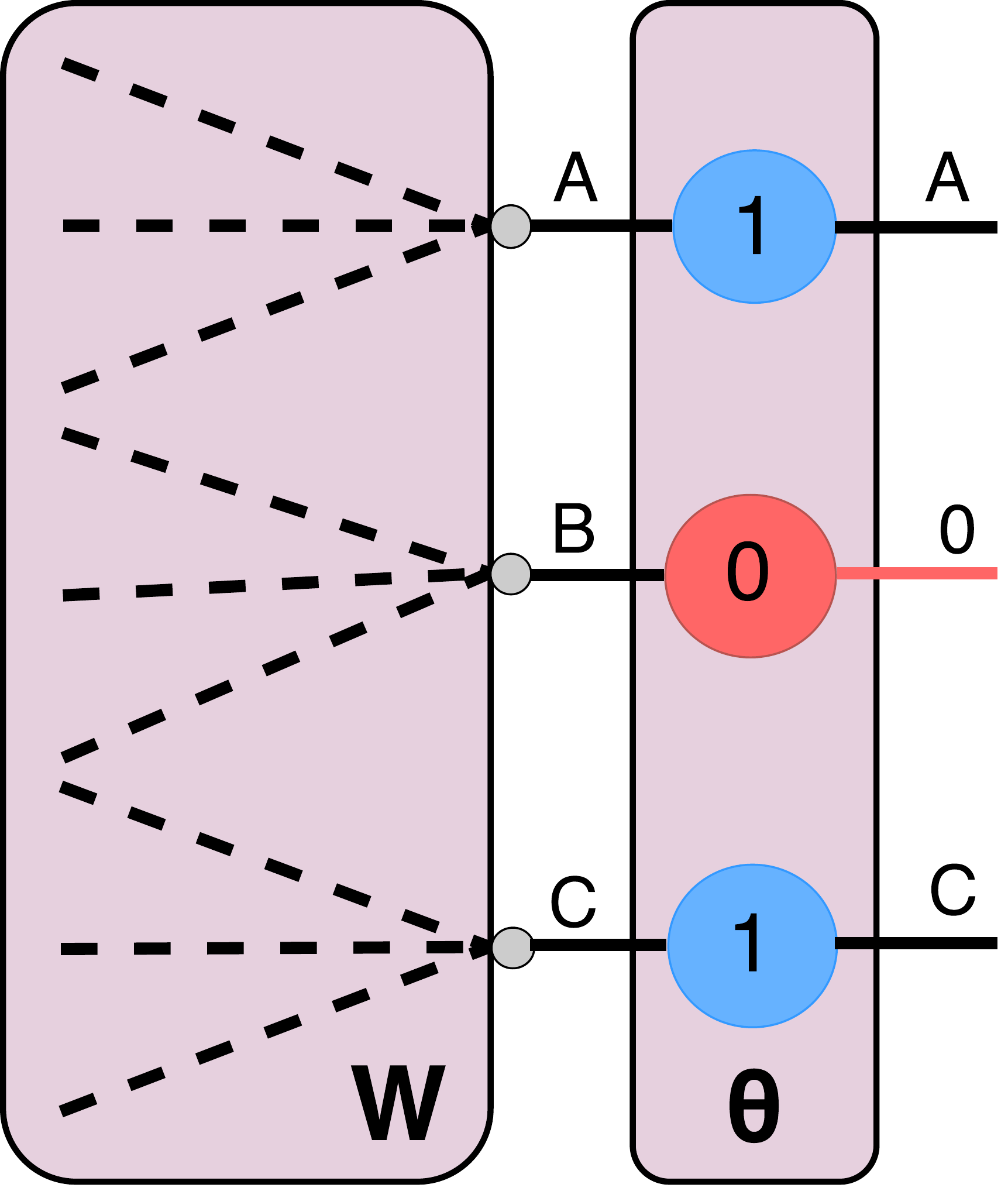}
\caption{{Neural Network Layer with Gates}}
\label{fig:nngates}
\end{subfigure} %
\begin{subfigure}{.5\textwidth}
  \centering
\includegraphics[width=7cm]{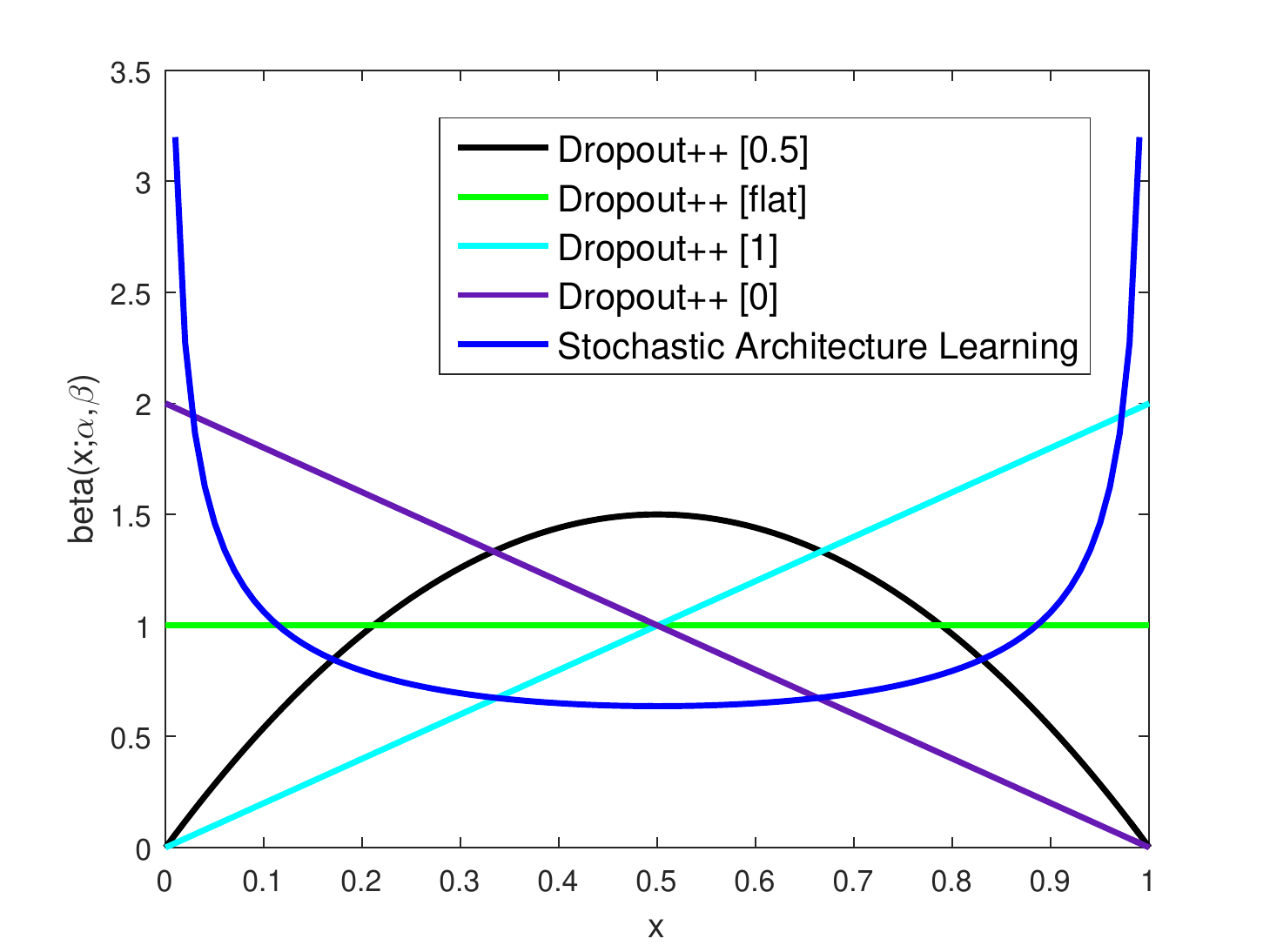}
\caption{{Beta distribution}}
\label{fig:betadist}
\end{subfigure}%
\caption{{\textbf{(a)} An illustration of the proposed method with binary stochastic multiplicative gates. Here, $W$ refers to weights and $\THT$ refers to gates. Note that $P(\THT = 1) = \K$.
\textbf{(b)} Behaviour of the beta distribution at different values of $\alpha, \beta$.}}
\end{figure*}

We add multiplicative gates after each feature / neuron in the neural network, as in Figure \ref{fig:nngates}. These gates modulate the output of each neuron. Let these new parameters be denoted by $\THT$. Let us also assume that they lie in $[0,1]$. Intuitively, these gates now control the \emph{relative importance} of each particular feature as viewed by the next layer. Such relative importance may be fundamentally uncertain given a particular feature. Hence, it may be useful to think of gate parameters $\THT$ as random variables. We shall now crystallize all these assumptions in the form of choices for the variational distribution.

We first place the following prior-hyperprior pair over the gate parameters $\THT$ , and a prior over the regular parameters $\W$. 

\begin{equation}
\THT  \sim  \prod_i bernoulli(k_i)  ~~~~~~
\K  \sim  \prod_i beta(\alpha,\beta)  ~~~~~~
\W  \sim \prod_i \mathcal{N}(0, \sigma)
\end{equation}

Note that the products are over all possible variables defined in the network. Here, $\K$ denotes the bernoulli parameters and also needs to be estimated along with $\THT, \W$. Now, given that we use variational inference, let us now define the forms of the variational distributions $q(\W,\THT,\K ; \A)$ we use. Let $\A = \{\A_1, \A_2\}$.

\begin{equation}
q(\W, \THT, \K; \A)  = q_1(\W; \A_1) ~q_2(\K;\A_2)~q_3(\THT; \A_2)  \nonumber\\
\end{equation}

\begin{equation}
q_1(\W)= \prod_i \delta(w_i ~;~ \mu_1^i)  ~~~~~~~~ q_2(\K) = \prod_i \delta(k_i ~;~ \mu_2^i) ~~~~~~~~ q_3(\THT) = \prod_i bernoulli(\theta_i ~;~ \mu_2^i) \label{WVariational}\\
\end{equation}

Note that even though we make an independence assumption on the weights $\W$ (equation \ref{WVariational}), we overcome the disadvantages described in the previous section by effectively \textbf{not} being Bayesian with respect to $\W$, using a delta distribution. 
 Also note that we use the same parameter $\A_2$ for both distributions $q_2(.)$ and $q_3(.)$. While it is true that using different parameters for both distributions could make the formulation more powerful, we use the same parameter for simplicity. Now we write equations describing the variational approximation by using the definitions above.

\begin{eqnarray}
\A_1^*, \A_2^*  = & \underset{\A_1, \A_2}{\arg\min} & ~~ \mathrm{KL}[q(\W , \THT, \K; \A) || P(\W, \THT, \K|\D)] \nonumber\\
= & \underset{\A_1, \A_2}{\arg\min} & ~~ -\sum_\theta \log P(\D|\THT,\W,\K)~q_3(\THT;\A_2)   - \log P(\K; \alpha, \beta) \label{ObjFun}
\end{eqnarray}

Our objective is now to solve equation \ref{ObjFun}. We observe that the exhaustive summation in equation \ref{ObjFun} in intractable for large models. A popular method to deal with this is to use a Monte-Carlo approximation of the summation. However, even this may be infeasible for large models. As a result, we further approximate this with a \emph{single} Monte-Carlo sample. In other words, we perform the following approximation:

\begin{eqnarray*}
\sum_\theta \log P(\D|\THT,\W,\K )~q_3(\THT;\A_2) & \approx & \log P(\D|\THT^s,\W,\K)\\
\mathrm{where}~~ \THT^s& \sim& q_3( \THT; \A_2) 
\end{eqnarray*}

While this approximation seems to be drastic, we soon shall see that Classic Dropout also implicitly performs the same approximation.

\subsection{Generalized Dropout}
Given all the assumptions and approximations discussed above, we now write the complete objective function we aim to solve. Since the variational distributions for $\W$ and $\K$ are delta distributions, we shall now use $\W,\K$ instead of $\A_1, \A_2$ in our notations, for simplicity.

\begin{eqnarray}
\W^*, \K^*  = & \underset{\W, \K}{\arg\min} & ~~ - \log P(\D|\THT^s,\W,\K)   -   (\alpha-1) \log \K  - (\beta - 1) \log (1 - \K) \label{ActualObjFun}\\
&\mathrm{where}&~~ \THT^s \sim bernoulli( \THT; \K) \nonumber 
\end{eqnarray}

In the expression above, we have used the fact that $P(\K ; \alpha, \beta)$ is a beta distribution. This form of the objective function \ref{ActualObjFun}, with gates $\THT,\K$ constitutes the \emph{Generalized Dropout} regularizer. 

Let us now briefly look at the behaviour of the beta distribution at various values of $\alpha, \beta$, as shown in Figure \ref{fig:betadist}. We shall refer to each of these specific cases as different versions of \emph{Dropout++}. For reasons to be discussed later, we shall refer to the last case as \emph{Stochastic Architecture Learning} (SAL).

\begin{itemize}
\item \textbf{Dropout++ (0.5)}, where $\alpha = \beta > 1$: $\K = 0.5$ is the most probable value of $\K$. 
\item \textbf{Dropout++ (flat)}, where $\alpha = \beta = 1$: All values of $\K$ are equally probable.
\item \textbf{Dropout++ (1)}, where $\alpha = 1, \beta > 1 $: $\K = 1$ is the most probable value of $\K$.
\item \textbf{Dropout++ (0)}, where $\alpha > 1, \beta = 1 $: $\K = 0$ is the most probable value of $\K$.
\item \textbf{SAL}, where $\alpha < 1, \beta < 1 $: $\K = 0$ and $\K = 1$ are the most probable values of $\K$.
\end{itemize}

Note that Dropout++ (0.5) becomes indistinguishable from Classic Dropout (with 0.5 Dropout rate) at $\alpha = \beta \rightarrow \infty$. To obtain other Dropout rates, we simply ensure $\alpha \neq \beta$. In the next section, we shall discuss another algorithm called \emph{Architecture Learning}, and how it relates to the \emph{SAL} method above.

\subsection{Architecture Learning}
Srinivas and Babu \cite{srinivas2015learning} recently introduced a method to learn the width and depth of neural network architectures. They also add additional learnable parameters similar to gates. Also, their objective function has the following form in our notation.

\begin{eqnarray}
\W^*, \K^*  = & \underset{\W, \K}{\arg\min} & ~~ - \log P(\D|\THT^s,\W,\K)   +   \lambda_1 \K (1 - \K) + \lambda_3 \K  \label{ALObjFun}\\
&\mathrm{where}&~~ \THT^s  = heaviside(\K - 0.5) \nonumber 
\end{eqnarray}

Note that our objective function \ref{ActualObjFun} looks very similar to this when $\alpha, \beta < 1$ and $\alpha < \beta$, except that we use $\log \K$ instead of $\K$. Another difference is that they use a heaviside threshold to select $\THT$ rather than sampling from a bernoulli. We observe that this is equivalent to taking a maximum likelihood sample from the bernoulli distribution. Given these similarities, we found it apt to name the corresponding method with $\alpha, \beta < 1 $ as \emph{Stochastic Architecture Learning}, as it is a stochastic version of the algorithm described above.

Most surprisingly, we find that the motivation to arrive at this algorithm was completely different - they intended to minimize the number of neurons in the network. We arrive at a very similar formulation from a purely Bayesian perspective. 

\subsection{A Practitioner's Perspective}
In this section, we shall attempt to provide an intuitive explanation for Generalized Dropout. Going back to Fig. \ref{fig:nngates}, each neuron is augmented with a gate which learns values between 0 and 1. This is enforced by our regularizers and well as by parameter clipping. During the forward pass, we treat each of these gate values as probabilities and toss a coin with that probability. The output of the coin toss is used to block / allow neuron outputs. As a result of the learning, important features tend to have higher probability values than unimportant features.

At test time, we do not perform any sampling. Rather, we simply use the real-valued probability values in the gate variables. This approximation - called re-scaling - is used in classical Dropout as well. 

What do the different Generalized Dropout methods do? Intuitively, they place restriction on the gate values (probabilities) that can be learnt. As an example, Dropout++ (0) encourages most gate values to be close to $0$, with only a few important ones being high. On the other hand, Dropout++ (1) encourages gates values to be close to $1$. Intuitively, this means that Dropout++ (0) restricts the capacity of a layer by a large amount, whereas Dropout++ (1) hardly changes anything. SAL, on the other hand, encourages neurons to be close to either $0$ or $1$. In contrast to other methods, SAL produces neural network layers that are very close to being deterministic - neurons close to $0$ are almost never 'on' and those close to $1$ are almost always 'on'. Dropout++ (flat) is also unique in the sense that it doesn't place any restriction on the gate values. As a result, we do not require to set any hyper-parameters for this method. From a Bayesian Perspective, when we have no prior beliefs on what the gate values should be, we use the most non-informative prior - which is Dropout++ (flat) in this case.

Dropout++ (0.5) encourages values to be close to 0.5. If the regularization constants are increased, then gate values other than 0.5 are penalized more and more heavily. In the limiting case we get Dropout, where any deviation from probability value of 0.5 is "infinitely" penalized.

\subsection{Estimating gradients for binary stochastic gates}
Given our formalism of stochastic gate variables, it is unclear how one might compute error gradients through them. Bengio \emph{et al.} \cite{bengio2013estimating} investigated this problem for binary stochastic neurons and empirically verified the efficacy of different solutions. They conclude that the simplest way of computing gradients - the \textit{straight-through} estimator works best overall. This involves simply back-propagating through a stochastic neuron as if it were an identity function. If the sampling step is given by $\theta \sim bernoulli(k)$, then the gradient $\frac{d\theta}{dk} = 1$ is used.

Another issue of consideration is that of ensuring that $k$ always lies in $[0,1]$ so that it is a valid bernoulli parameter. Bengio \emph{et al.} \cite{bengio2013estimating} use a sigmoid activation over $k$. Our experiments showed clipping functions worked better. This can be thought of as a `linearized' sigmoid. The clipping function is given by the following expression.

\begin{equation*}
clip(k) = \begin{cases}
1, & k \geq 1 \\
0, & k \leq 0 \\
k, & otherwise
\end{cases}
\end{equation*}

The overall sampling function is hence given by $\theta \sim bernoulli(clip(k))$, and the straight-through estimator is used to estimate gradients overall.

\subsection{Applying to Convolution Layers}
Here we shall discuss how to apply this to convolutional layers. Let us assume that the output feature map from a convolutional layer is $k \times k \times n$, i.e; $n$ feature maps of size $k \times k$. Classical dropout samples $k \times k \times n$ bernoulli random variables and performs pointwise multiplication with the output feature map. We follow the same for Generalized Dropout as well.

However, if we wish to perform architecture selection like Architecture Learning \cite{srinivas2015learning}, we need to select a subset of the $n$ feature maps. In this case, we only have $n$ gate variables, multiplying to the output of each feature map. When a gate is close to zero, and entire feature map's output becomes close to zero at test time. By selecting few feature maps out of $n$, we determine which of the $n$ filters in the previous layer are essential.

\section{Related Work}
There are plenty of works which aim to extend Dropout. DropConnect \cite{wan2013regularization} stochastically drops weights instead of neurons to obtain better accuracy on ensembles of networks. As stated earlier, using the independence assumption for weights may not be correct. Indeed, DropConnect is shown to work on only fully connected layers. Standout \cite{ba2013adaptive} is a version of Dropout where the dropout rate depends on the output activations of a layer. Variational Dropout \cite{kingma2015variational} proposes a Bayesian interpretation for Gaussian Dropout rather than the canonical multiplicative Dropout. By considering multiplicative Dropout, we make important connections to Architecture Learning / neuron pruning. Gal and Gharamani \cite{gal2015bayesian} showed a Bayesian interpretation for binary dropout and show that test performance improves by performing Monte-Carlo averaging rather than re-scaling. For simplicity, we use the re-scaling method at test time for Generalized Dropout. Our work can be seen as an extension of this work by considering a hyper-prior along with a bernoulli prior.

Hinton and Van Camp \cite{hinton1993keeping} first introduced variational inference for making Neural Networks Bayesian. Recent work by Graves \cite{graves2011practical} and Blundell et al. \cite{blundell2015weight} further investigated this notion by using different priors and relevant approximations for large networks. Probabalistic Backpropagation \cite{hernandez2015probabilistic} is an algorithm for inferring marginal posterior probabilities for special classes of Bayesian Neural Networks. Our method is different from any of these methods as they are all Bayesian over the weights, whereas we are only Bayesian with respect to the gates.

\begin{figure*}[!t]
\begin{subfigure}{.33\textwidth}
  \centering
\includegraphics[width=5cm]{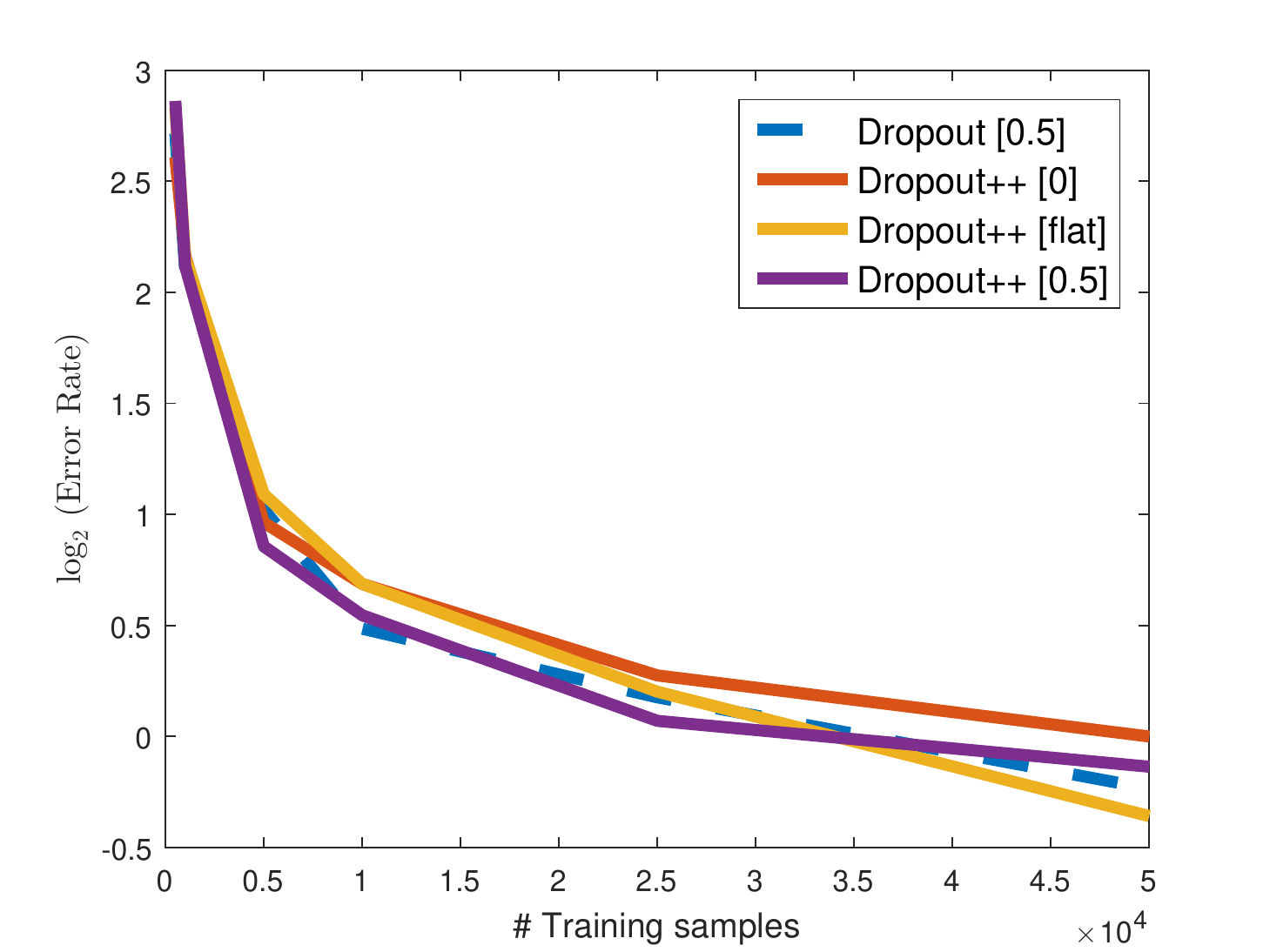}
\caption{\small{Effect of Large data}}
\label{fig:alldata}
\end{subfigure} %
\begin{subfigure}{.33\textwidth}
  \centering
\includegraphics[width=5cm]{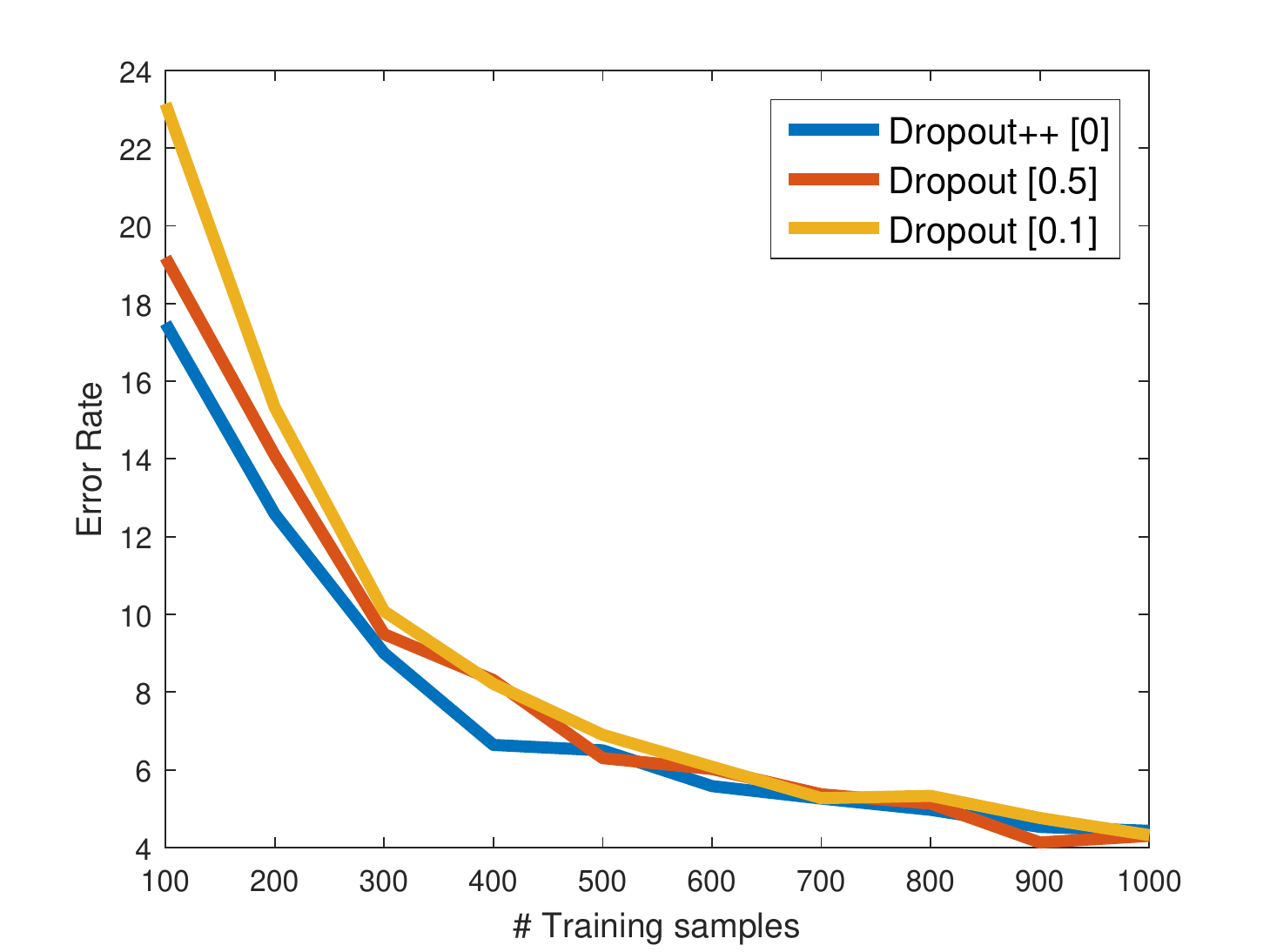}
\caption{\small{Effect of Small data}}
\label{fig:lowdata}
\end{subfigure}%
\begin{subfigure}{.33\textwidth}
  \centering
\includegraphics[width=5cm]{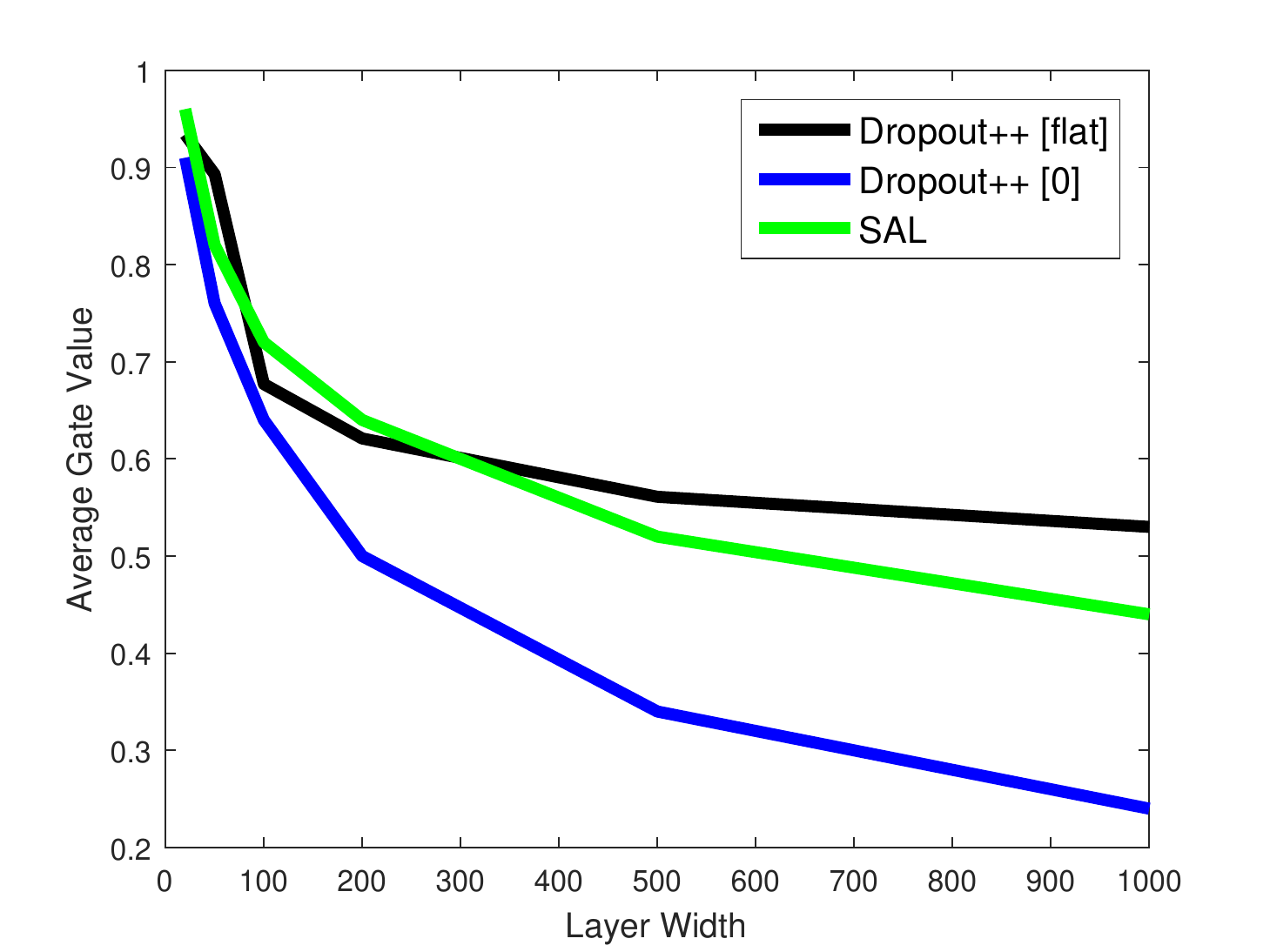}
\caption{\small{Gate value vs Layer Width}}
\label{fig:gatewidth}
\end{subfigure} %

\begin{subfigure}{.33\textwidth}
  \centering
\includegraphics[width=5cm]{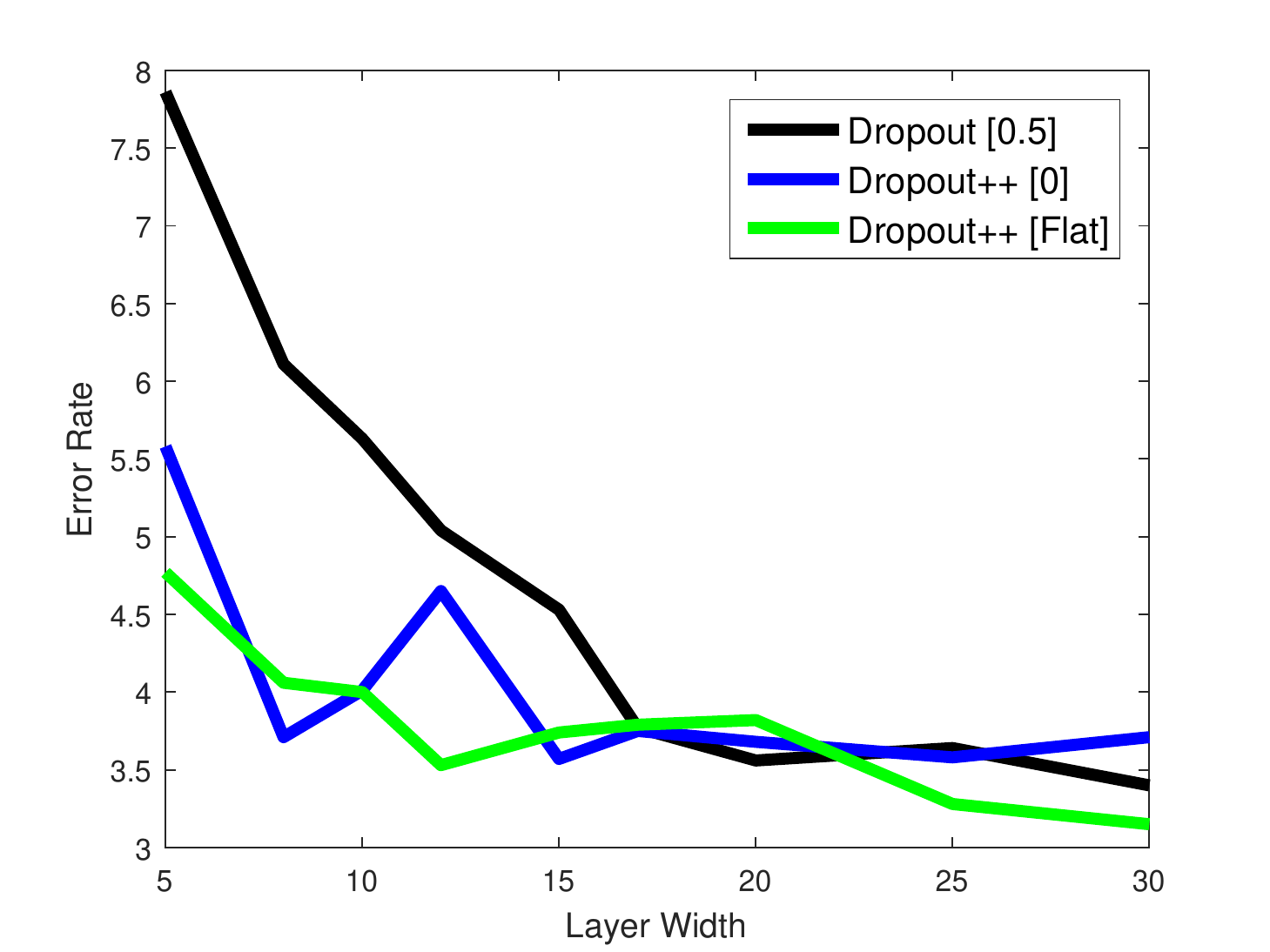}
\caption{\small{Accuracy vs Layer Width}}
\label{fig:width}
\end{subfigure}%
\begin{subfigure}{.33\textwidth}
  \centering
\includegraphics[width=5cm]{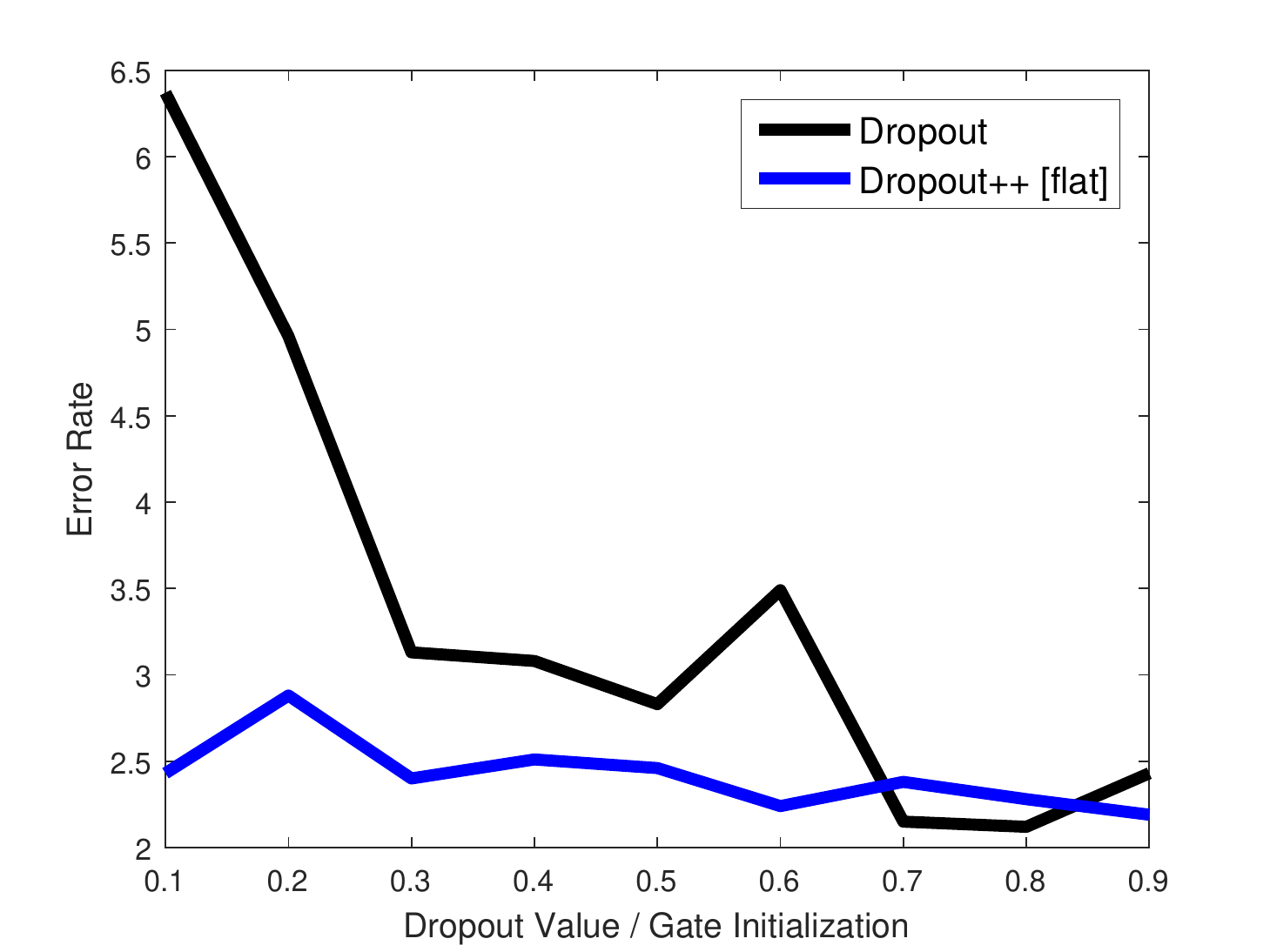}
\caption{\small{Effect of initialization}}
\label{fig:initialization}
\end{subfigure}%
\begin{subfigure}{.33\textwidth}
  \centering
\includegraphics[width=5cm]{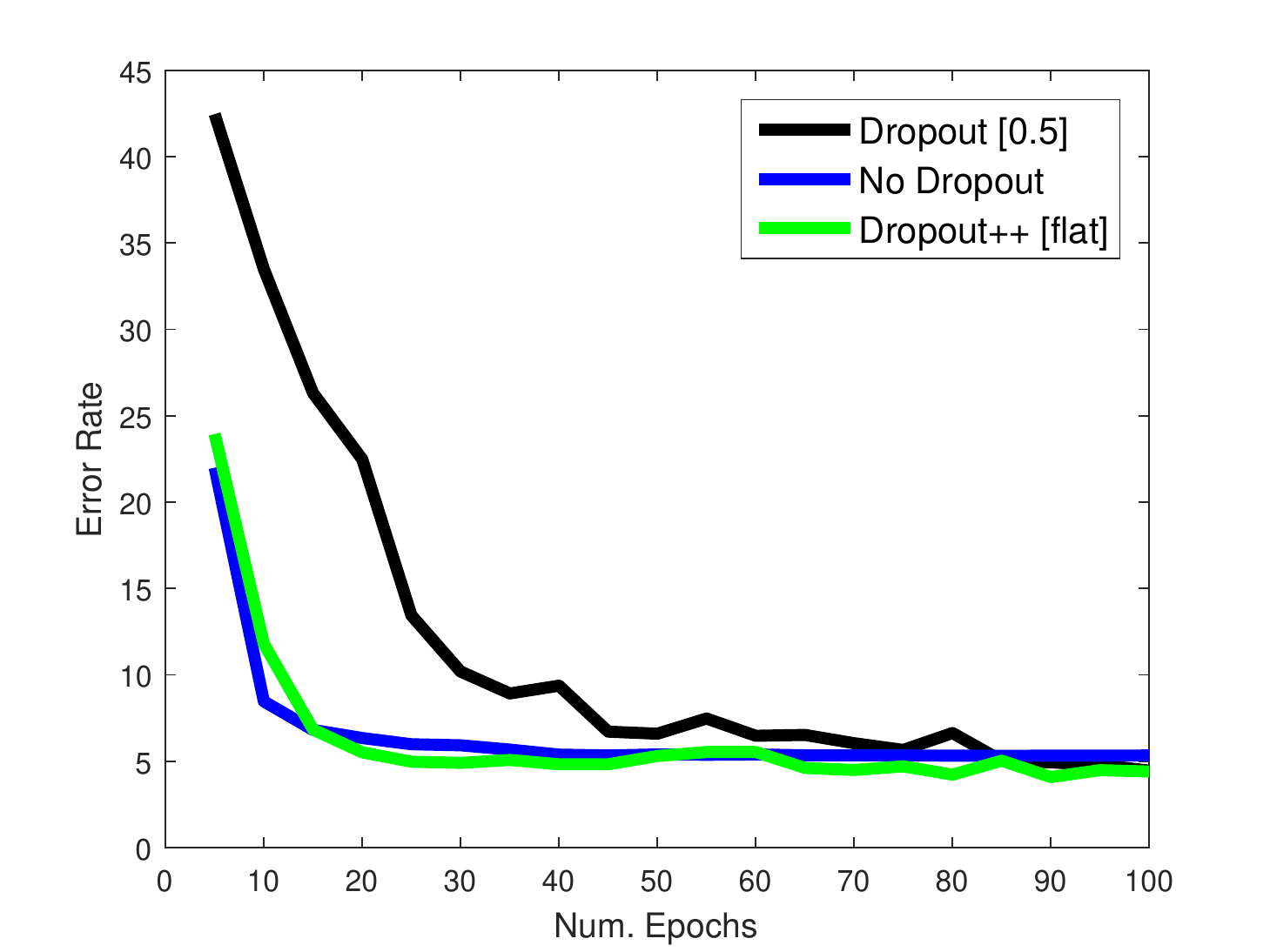}
\caption{\small{Speeding up Training}}
\label{fig:traintime}
\end{subfigure}%

\begin{subfigure}{.33\textwidth}
  \centering
\includegraphics[width=5cm]{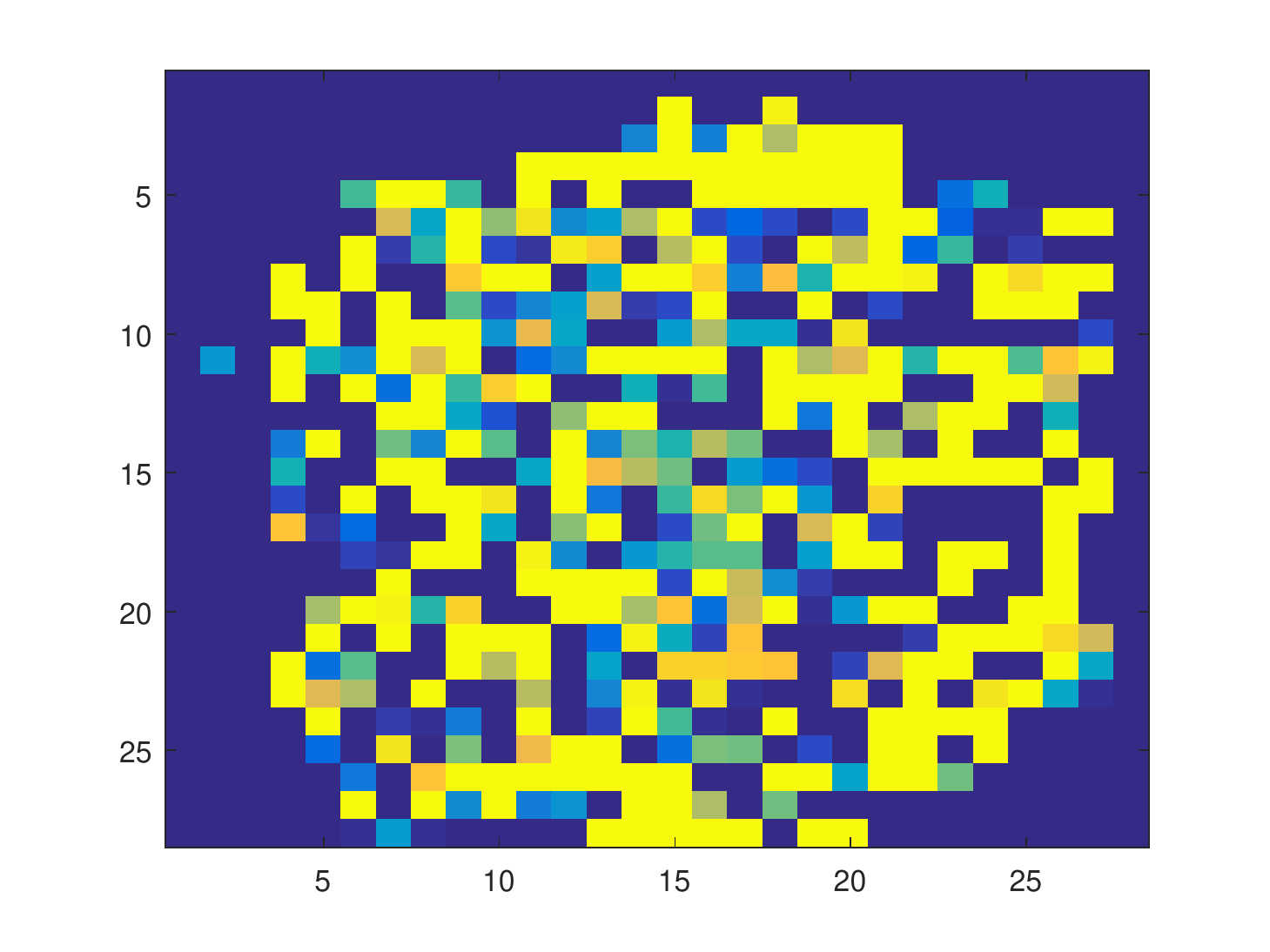}
\caption{\small{Input Layer Gate values}}
\label{fig:inputgateviz}
\end{subfigure}%
\begin{subfigure}{.33\textwidth}
  \centering
\includegraphics[width=5cm]{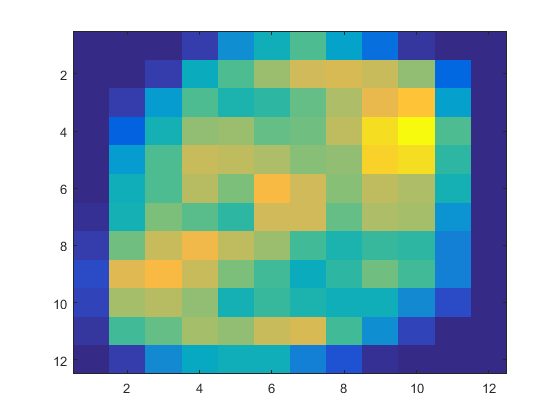}
\caption{\small{Conv. Layer (mean) Gate values}}
\label{fig:vizualization}
\end{subfigure}%
\begin{subfigure}{.33\textwidth}
  \centering
\includegraphics[width=5cm]{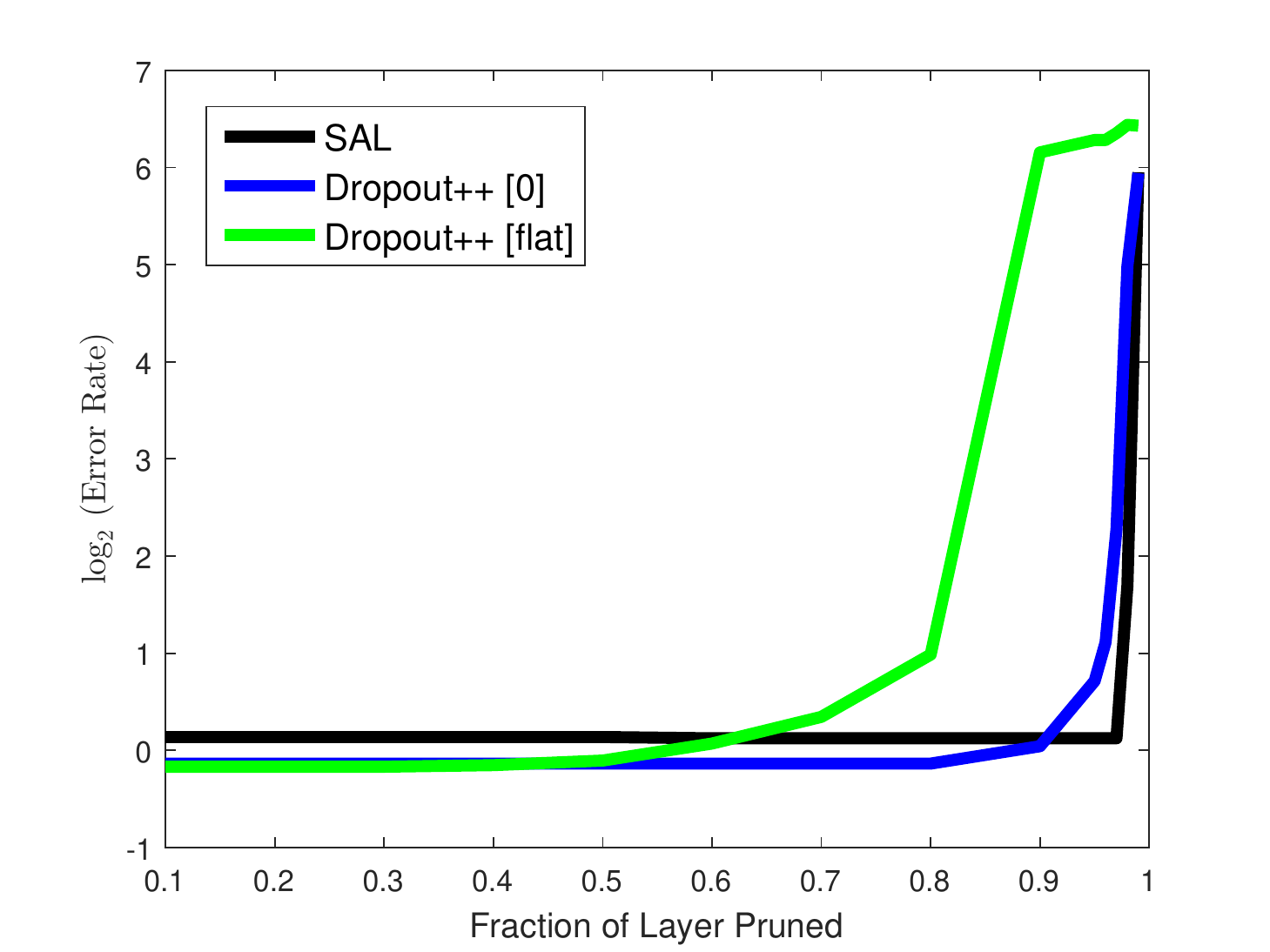}
\caption{\small{Pruning Gates}}
\label{fig:pruning}
\end{subfigure}%

\caption{{\textbf{(a, b)} All Dropout++ methods perform on par with Dropout on large data, and Dropout++ (0) seems to work better on small data. \textbf{(c, d)} Dropout++ can adapt to different layer sizes and result in optimal performance. \textbf{(e, f)} Initialization of Dropout++ parameters is not very crucial. As a result, one can initialize favourably to drastically increase training speed. \textbf{(g, h)} Dropout++ on convolutional layers learns to selectively attend to parts of the image, rather than the full image. \textbf{(i)} SAL and Dropout++ (0) are best suited for gate pruning, with SAL being better for automated pruning given the steep slope.}    }
\vspace{-0.5cm}
\end{figure*}

\section{Experiments}
In this section, we perform experiments with the Generalized Dropout family to test their usefulness. First, we perform a wide variety of analysis with the Generalized Dropout family. Later, we study some specific applications of this method. We perform experiments primarily using Theano \cite{bergstra2010theano} and Lasagne. 

\subsection{Analysis of Generalized Dropout}
We shall now analyze the behaviours of different members of Generalized Dropout family to find out which ones are useful. For the experiments on the MNIST dataset, we use the standard LeNet-like architecture \cite{lecun1998gradient}, which consists of two $5 \times 5$ convolutional layers with 20 and 50 filters, and two fully connected layers with 500 and 10 (output layer) neurons. While there is nothing particularly special about this architecture, we simply use this as a standard net to analyze our method.

\subsubsection{Effect of data-size}
We investigate whether Generalized Dropout indeed has any advantage over Dropout in terms of accuracy. Here, we apply Dropout and Generalized Dropout only to the last fully connected layer. Our experiments reveal that for the network considered, the accuracies achieved by any Generalized Dropout method are not always strictly better than Dropout, as shown in Figure \ref{fig:alldata}. This indicates that most of the \textit{regularization power} of Dropout comes from the independence assumption of Variational Inference, rather than particular values of the dropout parameter. This is a surprising result which we shall use to our advantage in the paper. 

However, we note that for small data-sizes, Dropout++ (0) seems to be advantageous over Dropout (Figure \ref{fig:lowdata}). This is possibly because Dropout++ (0) forces most neurons (but not all) to have very low capacity due to low value of the parameters. \footnote{Note that in our notation, a large value of Dropout++ indicates a large probability of retaining the neuron, contrary to popularly used notation for Dropout.}

\begin{table*}[!t]
\centering
\begin{tabular}{c|c|c|c}
\hline 
\textbf{Method}& \textbf{Architecture} & \textbf{Error (\%)} & \textbf{No. of Params} \\  
\hline \hline
Baseline & 20-50-500-10 & 0.82 & 431k\\
Architecture Learning \cite{srinivas2015learning} & 20-50-20-10 & 0.93 & 41.8k\\
SAL [$\beta / \alpha$ = 1] & 18-50-296-10 & 0.69 & 263k\\
SAL [$\beta / \alpha$ = 10] &  11-33-38-10 & 0.84 & \textbf{29.8k} \\
SAL [$\beta / \alpha$ = 100] & 7-13-16-10 & 1.14 & 5.9k \\
\hline 
\end{tabular} 
\caption{{Architecture selection capability of our method on a LeNet-like baseline. The first row is the architecture and performance of the baseline LeNet network. We see that larger $\beta / \alpha$ ratios result in smaller networks at the cost of network performance.}}
\label{table: mnist-archlearn}
\vspace{-0.5cm}
\end{table*}

\begin{table*}[!t]
\centering
\begin{tabular}{c|c||c|c||c|c}
\hline 
\textbf{\small{Model}}& \textbf{\small{Original}} & \textbf{\small{D++ ($init = 1.0$)}} & \textbf{\small{Drop ($p = 0.98$)}} & \textbf{\small{D++ ($init = 0.5$)}} & \textbf{\small{Drop ($p = 0.8$)}} \\  
\hline \hline 
ResNet-32 \cite{he2015deep} & 7.46 & \textbf{6.73} & 6.8 & - & - \\
ResNet-56 \cite{he2015deep} & 6.75 & \textbf{6.1} & 6.18 & - & - \\
GenericNet \cite{snoek2015scalable} & 6.86 & \textbf{6.6} & 6.75 & \textbf{6.32} & 6.46 \\
\hline 
\end{tabular} 
\caption{{Applying Dropout++ after each layer in standard networks decreases error rate (\%). Dropout++ learnt values (with init = 1.0) are close to $0.98$. As a result, Dropout at $p = 0.98$ performs similar to this learnt value. With Dropout++ init = 0.5, the learnt values are close to $p = 0.8$. For ResNets, small values of Dropout makes training difficult.}}
\label{table: gendrop-cifar}
\vspace{-0.5cm}
\end{table*}

\subsubsection{Effect of Layer-width}
Inspired from the above results about Dropout++ (0), we look at the relationship between using different layer-widths for the fully connected layer and the learnt gate parameters. Intuitively, it is natural to assume that larger layers should learn lower gate values, whereas smaller layers should learn much higher values, if we wish for the overall \textit{capacity} of the layer to remain roughly the same. Our experiments confirm this intuition as shown in Figure \ref{fig:gatewidth}. 

We also test if this flexibility translates to higher accuracy numbers over a fixed dropout value, and we find this to be indeed the case. We find that for small layer-widths, Dropout (at $p = 0.5$ for example), tends to remove too many neurons, while Dropout++ adjusts it's parameter values to account for small layer-widths, as shown in Figure \ref{fig:width}.

\subsubsection{Effect of Initialization}
Initialization of good parameters is known to play a key-role in generalization of deep learning systems. To test whether this holds for the newly introduced Generalized Dropout parameters as well, we try different initializations of the Generalized Dropout parameters. In this example, we simply initialize all gates to a single constant value. As expected, we find that the choice of this initialization is much less crucial when compared to setting the Dropout value, as shown in Figure \ref{fig:initialization}. 

The choice of initialization, however, affects training time. As an example, it is empirically observed that Dropout with $p = 0.1$ is much slower than $p = 0.9$. Therefore, it is helpful to have higher Dropout rates to facilitate faster training. To help faster training in Dropout++, we simply initialize with $p = 1.0$, i.e; start with a network with no Dropout and gradually learn how much Dropout to add. We observe that this indeed helps training time and at the same time provides the flexibility of Dropout, as shown in Figure \ref{fig:traintime}.

\subsubsection{Visualization of Learnt Parameters}
Until this point, we have focussed on using Generalized Dropout on the fully connected layers. Similar effects hold when we apply these to convolutional layers as well. Here, we visualize the learnt parameters in convolutional layers. First, we add Dropout++ only to the input layer. The resulting gate parameters are shown in Figure \ref{fig:inputgateviz}. We observe a similar effect when we add Dropout++ only to the first convolutional layer, as shown in Figure \ref{fig:vizualization}, which shows the average gate map of all the convolutional filters in that layer. In both cases, we observe that Dropout++ learns to selectively attend to the centre of the image rather than towards the corners.

This has multiple advantages. First, by not looking at the corners of each feature, we can potentially decrease model evaluation time. Second, this breaks translation equivariance implicit in convolutions, as in our case certain spatial locations are more important for a filter than others. This could be helpful when using CNNs for face images (for example), where a filter need not look for an "eye" everywhere in the image. Such locally connected layers have been previously used in works such as DeepFace \cite{taigman2014deepface}. Dropout++ could offer a more natural way to incorporate such an assumption.

\subsubsection{Architecture Selection}
We shall now attempt to use Stochastic Architecture Learning (SAL) to automatically learn the required layer width of the network. The inherent assumption here is that the initial architecture is over-complete, and that a sub-set of neurons is sufficient to get similar performance. We first learn the parameters of the network using SAL regularizer, later we prune neurons with low gates parameters. Figure \ref{fig:pruning} shows that SAL learns gate parameters that are often close to either 0 or 1, resulting in a much sharper rise compared to the other methods. We use this sharp rise as a criterion to select the width of a layer. We observe that varying the $\beta / \alpha$ parameter encourages the method to get smaller architectures, sometimes at the cost of accuracy, as shown in Table \ref{table: mnist-archlearn}.

\subsection{Dropout++ on standard models}
So far we have studied the various properties of Generalized Dropout by performing various experiments on LeNet. We shall now shift to larger networks to test the effectiveness of Dropout++. Modern networks mainly use dropout only in the fully connected layers, or simply not at all, owing to much powerful regularizers such as Batch Normalization. Here we shall take such networks, simply add Dropout++ (flat) after each layer, and see if we get an increase in accuracy. We perform experiments with ResNet32, ResNet56 and a Generic VGG-like network, all trained on the CIFAR-10 dataset. As in Table \ref{table: gendrop-cifar}, we see that for all three models, adding Dropout++ is largely helpful.

\section{Conclusion}
We have proposed \textit{Generalized Dropout}, a family of methods that generalize Dropout-like behaviour. One set of methods in this family, \textit{Dropout++}, is an adaptive version of Dropout. \textit{Stochastic Architecture Learning} is another set of methods that performs architecture selection. An uninformed choice of the Dropout parameter usually hurts performance. Dropout++ helps in setting a useful parameter value regardless of factors such as layer width and initialization. Experiments show that it is generally beneficial to simply add Dropout++ (flat) after every layer of a Deep Network. 

\bibliography{gendropout}

\begin{thebibliography}{10}

\bibitem{krizhevsky2012imagenet}
Alex Krizhevsky, Ilya Sutskever, and Geoffrey~E Hinton.
\newblock Imagenet classification with deep convolutional neural networks.
\newblock In {\em Advances in Neural Information Processing Systems}, pages
  1097--1105, 2012.

\bibitem{srivastava2014dropout}
Nitish Srivastava, Geoffrey Hinton, Alex Krizhevsky, Ilya Sutskever, and Ruslan
  Salakhutdinov.
\newblock Dropout: A simple way to prevent neural networks from overfitting.
\newblock {\em The Journal of Machine Learning Research}, 15(1):1929--1958,
  2014.

\bibitem{gal2015bayesian}
Yarin Gal and Zoubin Ghahramani.
\newblock Bayesian convolutional neural networks with bernoulli approximate
  variational inference.
\newblock {\em arXiv preprint arXiv:1506.02158}, 2015.

\bibitem{bishop2006pattern}
Christopher~M Bishop.
\newblock Pattern recognition.
\newblock {\em Machine Learning}, 2006.

\bibitem{denil2013predicting}
Misha Denil, Babak Shakibi, Laurent Dinh, Nando de~Freitas, et~al.
\newblock Predicting parameters in deep learning.
\newblock In {\em Advances in Neural Information Processing Systems}, pages
  2148--2156, 2013.

\bibitem{srinivas2015learning}
Suraj Srinivas and R~Venkatesh Babu.
\newblock Learning the architecture of deep neural networks.
\newblock {\em arXiv preprint arXiv:1511.05497}, 2015.

\bibitem{bengio2013estimating}
Yoshua Bengio, Nicholas L{\'e}onard, and Aaron Courville.
\newblock Estimating or propagating gradients through stochastic neurons for
  conditional computation.
\newblock {\em arXiv preprint arXiv:1308.3432}, 2013.

\bibitem{wan2013regularization}
Li~Wan, Matthew Zeiler, Sixin Zhang, Yann~L Cun, and Rob Fergus.
\newblock Regularization of neural networks using dropconnect.
\newblock In {\em Proceedings of the 30th International Conference on Machine
  Learning (ICML-13)}, pages 1058--1066, 2013.

\bibitem{ba2013adaptive}
Jimmy Ba and Brendan Frey.
\newblock Adaptive dropout for training deep neural networks.
\newblock In {\em Advances in Neural Information Processing Systems}, pages
  3084--3092, 2013.

\bibitem{kingma2015variational}
Diederik~P Kingma, Tim Salimans, and Max Welling.
\newblock Variational dropout and the local reparameterization trick.
\newblock {\em arXiv preprint arXiv:1506.02557}, 2015.

\bibitem{hinton1993keeping}
Geoffrey~E Hinton and Drew Van~Camp.
\newblock Keeping the neural networks simple by minimizing the description
  length of the weights.
\newblock In {\em Proceedings of the sixth annual conference on Computational
  learning theory}, pages 5--13. ACM, 1993.

\bibitem{graves2011practical}
Alex Graves.
\newblock Practical variational inference for neural networks.
\newblock In {\em Advances in Neural Information Processing Systems}, pages
  2348--2356, 2011.

\bibitem{blundell2015weight}
Charles Blundell, Julien Cornebise, Koray Kavukcuoglu, and Daan Wierstra.
\newblock Weight uncertainty in neural networks.
\newblock {\em arXiv preprint arXiv:1505.05424}, 2015.

\bibitem{hernandez2015probabilistic}
Jos{\'e}~Miguel Hern{\'a}ndez-Lobato and Ryan~P Adams.
\newblock Probabilistic backpropagation for scalable learning of bayesian
  neural networks.
\newblock {\em arXiv preprint arXiv:1502.05336}, 2015.

\bibitem{bergstra2010theano}
James Bergstra, Olivier Breuleux, Fr{\'e}d{\'e}ric Bastien, Pascal Lamblin,
  Razvan Pascanu, Guillaume Desjardins, Joseph Turian, David Warde-Farley, and
  Yoshua Bengio.
\newblock Theano: a cpu and gpu math expression compiler.
\newblock In {\em Proceedings of the Python for scientific computing conference
  (SciPy)}, volume~4, page~3. Austin, TX, 2010.

\bibitem{lecun1998gradient}
Yann LeCun, L{\'e}on Bottou, Yoshua Bengio, and Patrick Haffner.
\newblock Gradient-based learning applied to document recognition.
\newblock {\em Proceedings of the IEEE}, 86(11):2278--2324, 1998.

\bibitem{he2015deep}
Kaiming He, Xiangyu Zhang, Shaoqing Ren, and Jian Sun.
\newblock Deep residual learning for image recognition.
\newblock {\em arXiv preprint arXiv:1512.03385}, 2015.

\bibitem{snoek2015scalable}
Jasper Snoek, Oren Rippel, Kevin Swersky, Ryan Kiros, Nadathur Satish,
  Narayanan Sundaram, Md~Patwary, Mostofa Ali, Ryan~P Adams, et~al.
\newblock Scalable bayesian optimization using deep neural networks.
\newblock {\em arXiv preprint arXiv:1502.05700}, 2015.

\bibitem{taigman2014deepface}
Yaniv Taigman, Ming Yang, Marc'Aurelio Ranzato, and Lior Wolf.
\newblock Deepface: Closing the gap to human-level performance in face
  verification.
\newblock In {\em Proceedings of the IEEE Conference on Computer Vision and
  Pattern Recognition}, pages 1701--1708, 2014.

\end{thebibliography}
\bibliographystyle{unsrt}

\end{document}